# Solving the Food-Energy-Water Nexus Problem via Intelligent Optimization Algorithms

Qi Deng, Zheng Fan, Zhi Li, Xinna Pan, Qi Kang, *Senior Member, IEEE*, MengChu Zhou, *Fellow, IEEE*

*Abstract*—The application of evolutionary algorithms (EAs) to multi-objective optimization problems has been widespread. However, the EA research community has not paid much attention to large-scale multi-objective optimization problems arising from real-world applications. Especially, Food–Energy–Water systems are intricately linked among food, energy and water that impact each other. They usually involve a huge number of decision variables and many conflicting objectives to be optimized. Solving their related optimization problems is essentially important to sustain the high-quality life of human beings. Their solution space size expands exponentially with the number of decision variables. Searching in such a vast space is challenging because of such large numbers of decision variables and objective functions. In recent years, a number of large-scale many-objectives optimization evolutionary algorithms have been proposed. In this paper, we solve a Food–Energy–Water optimization problem by using the state-of-art intelligent optimization methods and compare their performance. Our results conclude that the algorithm based on an inverse model outperforms the others. This work should be highly useful for practitioners to select the most suitable method for their particular large-scale engineering optimization problems.

## I. Introduction

Artificial intelligence (AI) plays an important role in the field of control or system optimization. It provides novel methods and cutting-edge technology advancements to address complex control problems and enhance system performance [1]-[9]. Improving the performance, intelligence, and adaptability of systems with AI can help us overcome the drawbacks of conventional approaches, which is crucial to achieving multiple system optimization goals.

AI technologies have the potential to generate precise system models and optimal judgments from vast datasets, particularly with the application of techniques like deep learning and reinforcement learning. AI-based intelligent control systems are capable of autonomously adjusting control strategies in response to changes in the surrounding environment and system status. Two examples of multi-agent systems that heavily utilize AI techniques to maximize and coordinate the cooperation and efforts of swarms are intelligent transportation systems and autonomous cars. AI is able to effectively deal with control problems in complex environments. In general, AI has progressively permeated all facets of manufacturing, control, and system optimization in today's era of rapid technological growth, emerging as a significant force in advancing social progress.

In the realm of AI, a number of intelligent optimization algorithms have been developed. Examples include simulated annealing, genetic algorithms, and particle swarm algorithms. These algorithms can be applied to control systems to address issues with resource allocation, path planning, and parameter optimization in order to boost system performance. With several competing objectives, a large number of application issues can be described as multi-objective optimization problems (MOPs). MOP is formally defined as:

$$\min F(x) = (f_1(x), f_2(x), \ldots, f_M(x))$$
$$\text{subject to } x \in \Omega \quad (1)$$

where $x = (x_1, x_2, \ldots, x_D)$ is a decision vector, $F(x)$ is an objective vector, $D$ and $M$ are dimensionalities of $x$ and $F(x)$, respectively. Multi-objective evolutionary algorithms (MOEAs), a sort of stochastic metaheuristic optimization techniques, have gained popularity for handling MOPs [10]-[15]. MOEAs are population-based search techniques that can be improved iteratively by reproducing new solutions from the current ones (called population) and updating them with the relatively best solutions till the present generation. Because of their population-based nature, EAs have the advantage of directing the optimization process toward the Pareto front (a set of optimal solutions) in multi-objective optimization problems.

When $M$ is greater than 3 and $D$ is greater than 500, (1) can be called large-scale many-objective problems (LSMOPs). When dealing with LSMOPs, multi-objective optimization algorithms suffer from poor spatial search efficiency because of the huge number of decision variables and complex objective functions. It is critical to 1) avoid blind (often useless) large-scale searches and 2) promote fast population convergence to the global optima more effectively. Therefore, a number of large-scale many-objective evolutionary algorithms have been proposed to solve this problem, and these MOEAs include FLEA [16], LERD [17], EAGO [18].

However, the majority have been assessed on the common benchmark test problems only [19], with relatively few algorithms being evaluated on real-world LSMaOPs.

This work for the first time applies an inverse model-based algorithm to solve the real-world large-scale many-objective

Q. Deng, Z. Fan, Z. Li and Q. Kang are with the Department of Control Science and Engineering, Tongji University, Shanghai 201804, China (e-mail: 1652365@tongji.edu.cn, 2180139@tongji.edu.cn, 2380124@tongji.edu.cn, qkang@tongji.edu.cn).
X. Pan is with Department of Trust & Safety of YouTube, Google, Mountain View, USA (e-mail: xinnapan@google.com).
M. Zhou is with ECE Department, New Jersey Institute of Technology, Newark, NJ 07102, USA (e-mail: zhou@njit.edu).

problem (Food–Energy–Water problems) and compare its outcomes with those of the other state-of-the-art ones.

## II. Food–Energy–Water Problem

The systems of food, energy, and water are highly complex. Planning human community systems, developing government policies, and making economic decisions all depend on quantifying the relationships and interconnections among food, energy, and water in such systems. Its optimal design and operation problems involve a huge number of variables and many objectives, thus falling into the class of large-scale many-objective optimization problems.

The numbers of water resources, energy sources, and food resources, and various demands for resources are represented by $\alpha$, $\beta$, $\gamma$, and $\varepsilon$ respectively. Food–Energy–Water Nexus [19]-[21] is established based on input–output theory:

$$x_j = \sum_{k=1}^{\alpha} x_j x_k + \sum_{k=1}^{\beta} x_j y_k + \sum_{k=1}^{\gamma} x_j z_k + \sum_{k=1}^{\varepsilon} x_j v_k \quad (2)$$

where $j = 1, 2, \ldots, \alpha$, $x_j x_k$, $x_j y_k$, $x_j z_k$ and $x_j v_k$ mean the amount of water $j$ used to generate water $k$, energy $k$, food $k$, and the amount of water $j$ used by demand $k$., respectively. $x_j$ is the consumption of water $j$. Similarly,

$$y_l = \sum_{k=1}^{\alpha} y_l x_k + \sum_{k=1}^{\beta} y_l y_k + \sum_{k=1}^{\gamma} y_l z_k + \sum_{k=1}^{\varepsilon} y_l v_k \quad (3)$$

$$z_i = \sum_{k=1}^{\alpha} z_i x_k + \sum_{k=1}^{\beta} z_i y_k + \sum_{k=1}^{\gamma} z_i z_k + \sum_{k=1}^{\varepsilon} z_i v_k \quad (4)$$

where $l = 1, 2, \ldots, \beta$ and $i = 1, 2, \ldots, \gamma$, $y_l$ and $z_i$ are the consumptions of energy $l$ and food $i$, respectively.

$$D = (\alpha + \beta + \gamma)(\alpha + \beta + \gamma + \varepsilon) \quad (5)$$

where $D$ is the number of decision variables of Food–Energy–Water Nexus. The consumptions of food, energy, and water are denoted by $x$, $y$ and $z$, respectively, i.e.,

$$x = \sum_{j=1}^{\alpha} x_j$$
$$y = \sum_{l=1}^{\beta} y_l$$
$$z = \sum_{i=1}^{\gamma} z_i$$
(6)

The quantity of water utilized to generate energy, water utilized to generate food, food utilized to generate water, energy utilized to generate food, and food utilized to generate energy are all included in the resource consumption and represented by $\Omega_{xy}$, $\Omega_{xz}$, $\Omega_{zx}$, $\Omega_{yz}$ and $\Omega_{zy}$, respectively.

$$\Omega_{xy} = \sum_{j=1}^{\alpha} \sum_{l=1}^{\beta} x_j y_l$$

$$\Omega_{xz} = \sum_{j=1}^{\alpha} \sum_{i=1}^{\gamma} x_j z_i$$

$$\Omega_{zx} = \sum_{i=1}^{\gamma} \sum_{j=1}^{\alpha} z_i x_j$$

$$\Omega_{yz} = \sum_{l=1}^{\beta} \sum_{i=1}^{\gamma} y_l z_i$$

$$\Omega_{zy} = \sum_{i=1}^{\gamma} \sum_{l=1}^{\beta} z_i y_l$$
(7)

Then we have the following optimization objectives:
$$\min f_1 = \Omega_{xy}/y$$
$$\min f_2 = \Omega_{xz}/z$$
$$\min f_3 = \Omega_{zx}/x$$
$$\min f_4 = \Omega_{yz}/z$$
$$\min f_5 = \Omega_{zy}/y$$
(8)

By maximizing resource production and decreasing resource consumption, the FEWN problem can be solved. A decrease in resource consumption and an increase in resource production are implied by minimizing the intensity coefficients: The idea of a low resource intensity is to use less to generate more. So $f_1$ means using less water to generate more energy and other objectives has similar explanations ($f_2$: water for food, $f_3$: energy for water, $f_4$: energy for food and $f_5$: food for energy).

## III. State-of-the-art Evolutionary Algorithms

Recently, some large-scale and many-objective algorithms [16]-[18] have been proposed and they can achieve great performance in solving some LSMaOPs.

### A. Fast Large-Scale MOEA Framework with Reference-Guided Offspring Generation (FLEA)

FLEA [16] is a large-scale MOEA framework that employs reference vectors to aid in the selection of promising solutions during a solution generation process, as shown in Fig. 1. Many examples of bidirectional convergence and diversity vectors that produce diverse search behaviors in decision space are given and used. Computational resources are allocated appropriately to balance convergence and diversity. FLEA's convergence direction vector uses local distribution information to drive quick searches in high-dimensional decision space. Its diversity direction vectors are used to generate a broader range of viable non-dominant solutions to promote population diversity.

FLEA can significantly increase the performance of large-scale searches without necessitating expensive problem reformulation or decision variable analysis procedures.

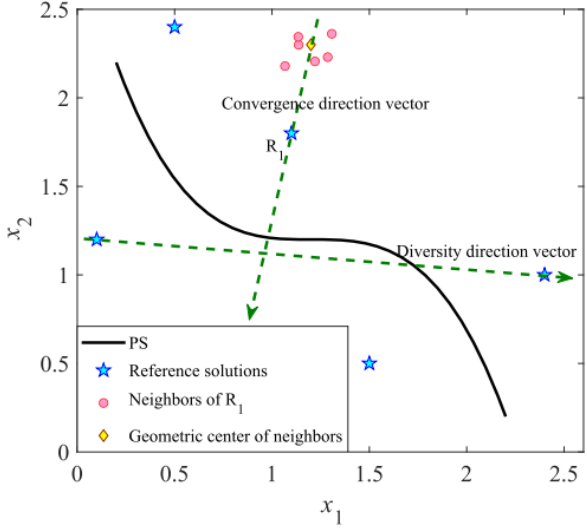

Fig. 1. Searching solutions in decision space assisted with reference vectors.

### B. MOEA Based on Reformulated Decision Variable Analysis (LERD)

LERD [17] uses an efficient reconstruction-based decision variable analysis (DVA) method as its core feature is summarized next, where *n* is the population size.

1. Initialize a population with *n* solutions.
2. Realize a binary vector-based decision variable analysis process.
3. Optimize convergence-related and diversity-related subproblems separately.
4. Develop an environmental selection approach to achieve the ultimate optimal population P.

DVA is restated as an optimization problem by using binary choice variables, with the goal of approximating distinct grouping results and dynamically estimating variable interactions in an evolutionary process to drive the group to the global optima.

### C. Evolutionary Algorithm with Population Generation via Objective Space (EAGO)

Different from existing studies, e.g., [16][17], EAGO [18] wants to collaboratively identify a novel solution by relating decision space and objective space, as shown in Fig. 2. In other words, it forms a cycle between decision space and objective space. The information from the objective space is now used to help generate new solutions in the form of feedback, hoping to significantly increase the algorithm's solution efficiency.

Unfortunately, achieving an ideal cycle is exceedingly challenging. EAGO chooses to set an inverse model as shown in Fig. 3. This algorithm has broken through the expensive and ineffective optimization paradigm of massive search in decision space by converting the conventional one-way search in decision space into a low-dimensional two-way-loop-optimization, by using the Pareto frontier neighborhood in objective space.

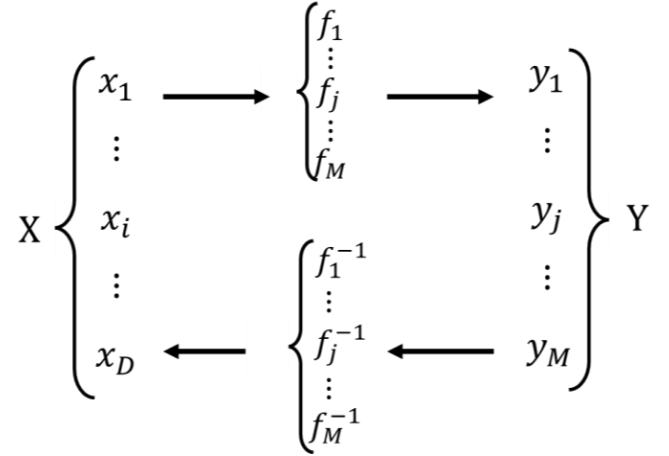

Fig. 2. Ideal optimization cycle.

EAGO builds a decision variable identification technique via objective space mapping. This strategy involves perturbing each decision variable in order to develop a specific population. Following space mapping, the general characteristic distribution of this population is used to determine the characteristics of decision variables. Different optimization methods are used to determine decision variables of different characteristics. Stated differently, during convergence optimization process, the variables related to diversity are maintained, allowing the population to efficiently iterate toward the Ideal Point; on the other hand, during diversity optimization process, the variables related to convergence are maintained, resulting in a more diverse population distribution.

In addition to minimizing the challenges involved in creating an ideal solution, this kind of decomposition can optimize subproblems of different characteristics, eliminating the need for some degree of blind large-scale searches. The population as a whole can converge due to the efficient identification of decision variables. As a result, the "curse of dimensionality" has less adverse effect on meta-heuristic algorithms because both convergence and diversity are ensured.

Since FLEA (2022) [16], LERD (2024) [17] and EAGO [18] are the latest large-scale and many objective algorithms and they will be applied to solve the Food–Energy–Water Nexus (FEWN) problem.

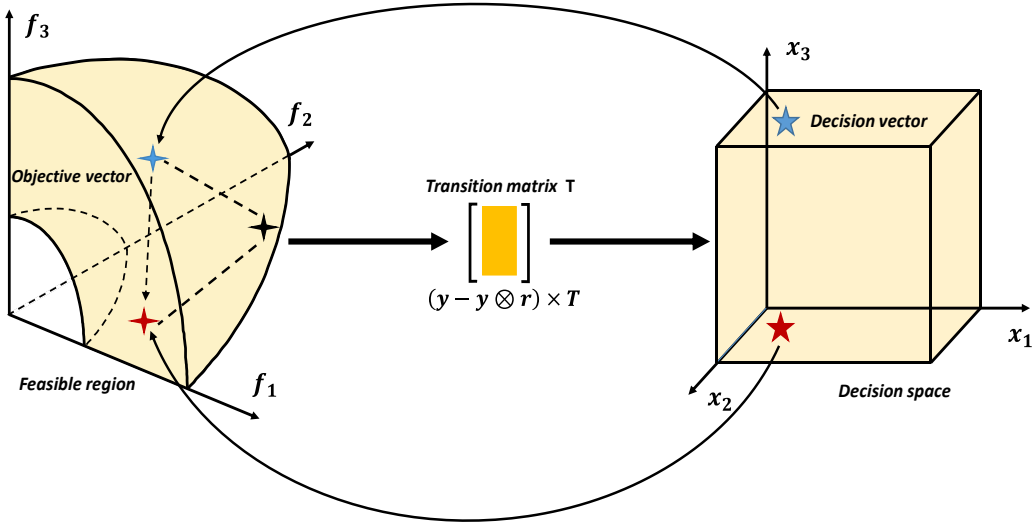

Fig. 3. Linear inverse model.

## IV. Solving FEWN Problems

The Food–Energy–Water Nexus (FEWN) problem encompasses the following resources: food (cereals, tubers, vegetables, fruits, and food products), energy (petroleum, hydro, wind, solar, and biofuels), and water (surface water, groundwater, desalinated water, wastewater reuse, and recycled water reuse). It also includes the final demands for these resources from households, the government, the rest of the economy, losses, storage, and exports. We can increase the number of resources in each category from 5 to 7 and obtain 567 variables according to (5). So this FEWN problem has 5 objectives and 567 decision variables.

We apply EAGO [18] to FEWN, along with two other state-of-the-art algorithms that are intended for large-scale many-objective optimization problems: FLEA (2022) [16] and LERD (2024) [17]. Since the true Pareto front of this problem is unknown, we utilize the Hypervolume (HV) index to gauge the results. Larger HV value indicates better performance.

$$\mathrm{HV} = \lambda(\cup_{i=1}^{|P|} x_i) \qquad (9)$$

where $\lambda$ is Lebesgue measure, $x_i$ is the hypervolume created by the reference point and a non-dominated point, and P is the non-dominated solution set.

The results on the FEWN problem are shown in Table I. From Table I, we can see that EAGO has an advantage in performance. Non-dominated solutions obtained by EAGO are shown in Fig. 4. we can design a more efficient FEWN system by using these solutions and make better policies to promote sustainable consumption and production of resources in human livelihood.

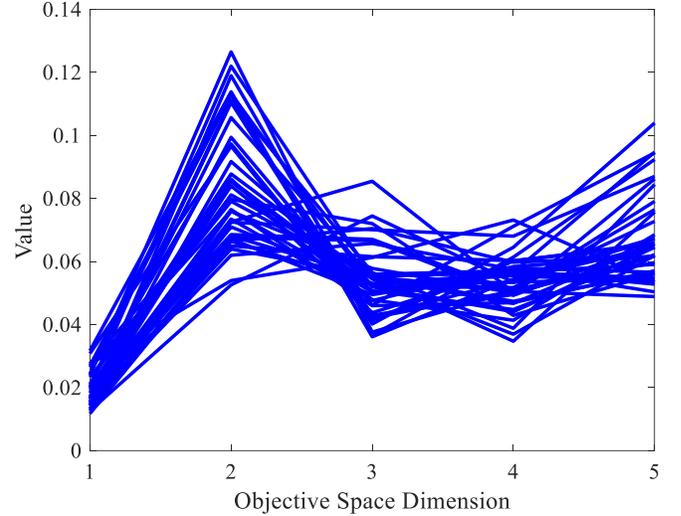

Fig. 4. The non-dominated solutions obtained by EAGO on FEWN problem.

## V. Conclusions

Comprehending various methods for reducing FEWN resource intensity is essential for formulating desired policies. An efficient system is one that uses less resources to generate more, or low resource intensity. We have applied an inverse model-based evolutionary algorithm to an FEWAN problem and have carried out a comparison analysis to assess how well SOTA algorithms work in resolving practical optimization problems. Future work aims to perform the best policy formation such that the FEWN resources can be intelligently allocated to generate the maximum benefits to human life.

TABLE I
HV VALUES OF THREE ALGORITHMS ON VEHICLESAFETY PROBLEM, WHERE BEST RESULTS ON EACH TEST INSTANCE ARE IN BOLD

| Problem | M | D | LERD | FLEA | EAGO |
|---|---|---|---|---|---|
| FEWN | 5 | 567 | 6.0296E-01 (6.37E-02) − | 5.6842E-01 (5.10E-02) − | **6.6219E-01 (7.85E-02)** |
| +/−/≈ | | | 0/1/0 | 0/1/0 | / |

The HV value is normalized to a range of 0-1, with larger values being better.